\documentclass[11pt,twoside,twocolumn,a4paper]{article}

\usepackage{cvww}
\usepackage{times}
\usepackage{epsfig}
\usepackage{graphicx}
\usepackage{amsmath}
\usepackage{amssymb}
\usepackage{enumitem}

\usepackage[pagebackref=true,breaklinks=true,bookmarks=false]{hyperref}
\cvwwfinalcopy % *** Uncomment this line for the final submission

\def\cvwwPaperID{6} % *** Enter the CVWW Paper ID here

\ifcvwwfinal\fi
\begin{document}

%%%%%%%%% TITLE
\title{Pose and Facial Expression Transfer by using StyleGAN}

\author{Petr Jahoda, Jan Cech\\
Faculty of Electrical Engineering, \\
Czech Technical University in Prague \\
% \and
% Third Author\\
% {\small\url{http://www.author.org/~third}}
}

\maketitle
\ifcvwwfinal\thispagestyle{fancy}\fi

%%%%%%%%% ABSTRACT
\begin{abstract}
We propose a method to transfer pose and expression between face
images. Given a source and target face portrait, the model produces an output image in which the pose and expression of the source face image are transferred onto the target identity. The architecture consists of
two encoders and a mapping network that projects the two inputs into the latent
space of StyleGAN2, which finally generates the output. The training is
self-supervised from video sequences of many individuals. Manual labeling is not required. Our model enables the synthesis of random identities with controllable pose and expression. Close-to-real-time performance is achieved.
\end{abstract}

%%%%%%%%% BODY TEXT
\section{Introduction}
Animating facial portraits in a realistic and controllable way has numerous
applications in image editing and interactive systems. For instance, a photorealistic
animation of an on-screen character performing various human poses
and expressions driven by a video of another actor can enhance the user experience in games or virtual reality
applications. Achieving this goal is challenging, as it requires representing the face (e.g. modeling in 3D) in order to control it and developing a method to map the desired form of control back onto the face representation.
%The form of control can be another face portrait. 

With the advent of generative models, it has become increasingly easier to
generate high-resolution human faces that are virtually indistinguishable from
real images. StyleGAN2 \cite{karras2020analyzing} achieves the state-of-the-art level of image generation with high quality and diversity among GANs \cite{NIPS2014_5ca3e9b1}.
% StyleGAN2 generates human faces by inputting a latent code, which is
% a vector sampled usually from Gaussian distribution, to the generator. We
% can semantically edit the images in the latent space, enabling us to change
% age, gender, smile, and other features. One common technique to do that
% is to identify linear semantic directions in the latent space and edit images
% by manipulating the latent code in these directions. However, these linear
% semantic directions are entangled, resulting in unwanted secondary edits (e.g.
% generating a person from a different viewpoint might make them grow a beard,
% age, change hairstyle, or change identity completely).
% Nevertheless, the generated images are still random, and we want to edit
% images of real people. GAN inversion aims to reconstruct an image of a real
% person by finding a latent vector that best represents the target image when
% sent through the generator. When the corresponding latent code is found, the
% aforementioned method can be used for editing. However, it still suffers from
% the same shortcomings.
Although extensive research has been conducted on editing images in the latent space of StyleGANs, most studies have primarily explored linear editing approaches. StyleGAN is popular for latent space manipulation using learned semantic directions, e.g. making a person smile, aging, change of gender or pose. However, the exploration of non-linear editing methods and example-based control of the synthesis remains relatively unexplored.

\begin{figure}[t]
    \centering
    \setlength{\tabcolsep}{0.005\linewidth}
    \renewcommand{\arraystretch}{0.03}
    \begin{tabular}{ccc}
        \includegraphics[width=0.32\linewidth]{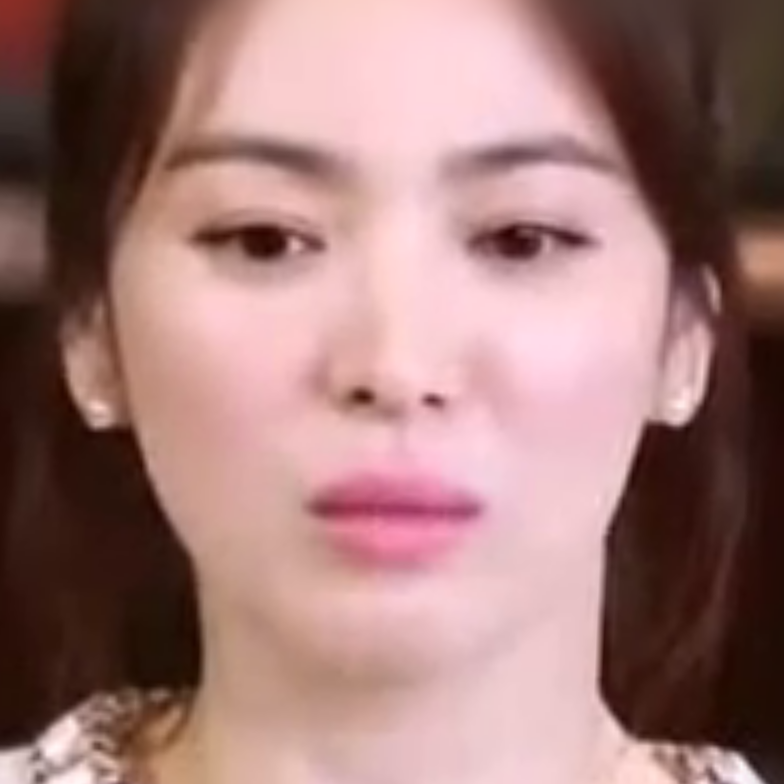} & 
        \includegraphics[width=0.32\linewidth]{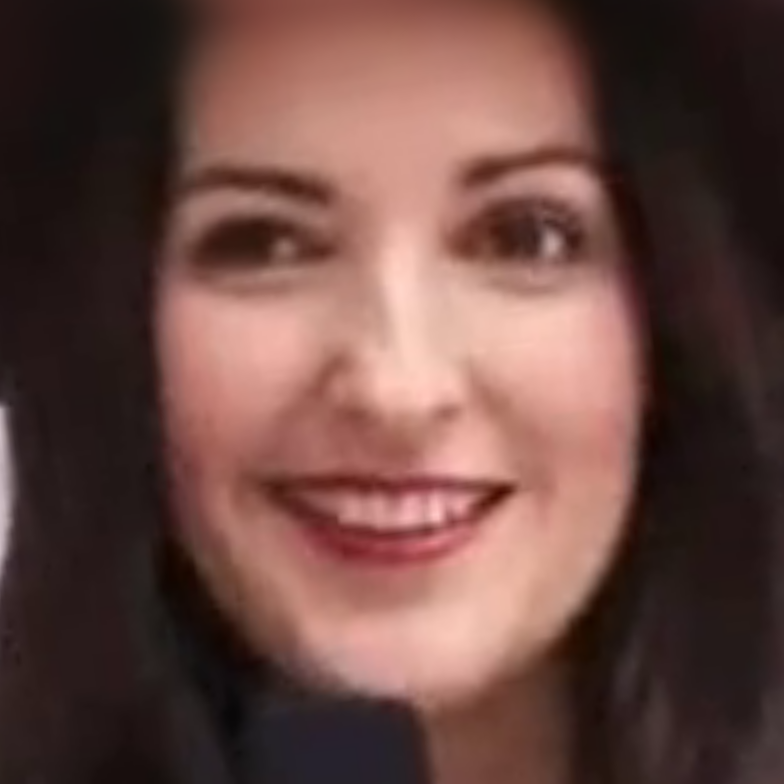} &
        \includegraphics[width=0.32\linewidth]{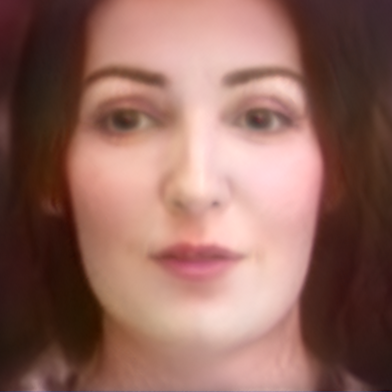} \\
        \vspace{0.01\textwidth} \\
        \includegraphics[width=0.32\linewidth]{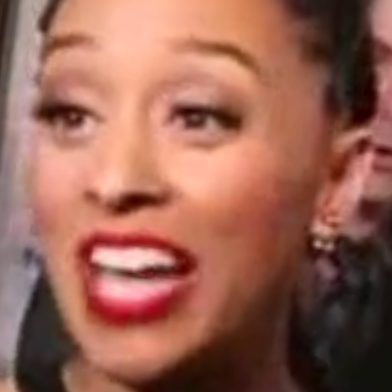} &
        \includegraphics[width=0.32\linewidth]{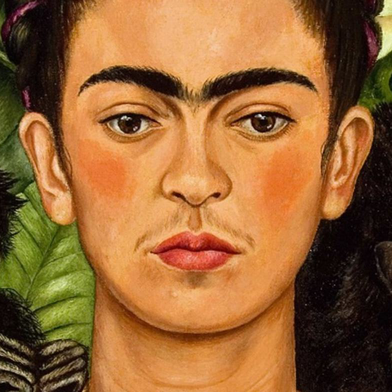} &
        \includegraphics[width=0.32\linewidth]{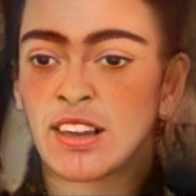} \\ 
        \vspace{0.01\textwidth} \\
        \includegraphics[width=0.32\linewidth]{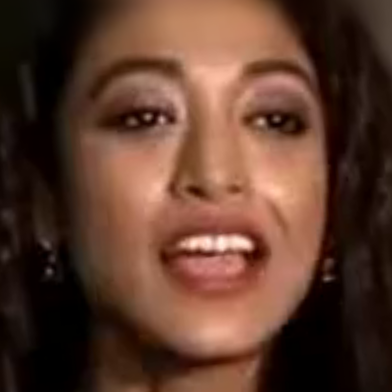} & 
        \includegraphics[width=0.32\linewidth]{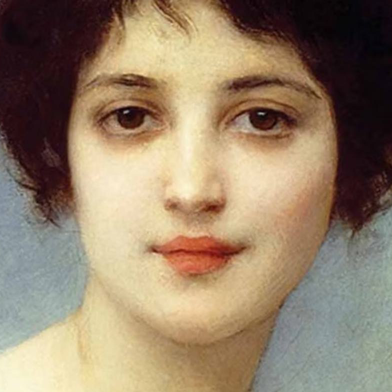} &
        \includegraphics[width=0.32\linewidth]{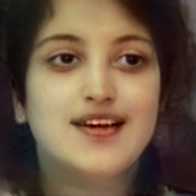} \\
    
        \vspace{0.01\textwidth} \\
        \raisebox{0.9cm}{Source} & 
        \raisebox{0.9cm}{Target} & 
        \raisebox{0.9cm}{Generated} \\[-3ex]
    \end{tabular}
    \caption{Results of our method. Pose and expression from the source image is transferred onto the identity of the target image. The method generalizes to paintings, despite being trained on videos of real people.}
    \label{fig:intro}
    \vspace*{-3mm}
    
\end{figure}

This work presents a method that synthesizes a new
image of an individual by taking a source (driving) image and a target (identity) image
as input, incorporating the pose and expression of the person in the source
image into the generated output from the target image, as shown in Fig.~\ref{fig:intro}.

%doplnit zakladni myslenku metody
% - pouziti predtrenovanych modelu styleganu2 k finalnimu generovani obrazu a restyle encoderu 
% - ucime se jen expression encoder a mapovaci sit
% - uceni z videi, self-supervised
% - jednoducha architektura, slozena z predtrenovanych bloku. 
% - vysledky ukazuji vysokou flexibilitu styleganu
The main idea of our method is to encode both images into pose/expression and identity embeddings. The embeddings are then mapped into the latent space of the pre-trained StyleGAN2~\cite{karras2020analyzing} decoder that generates the final output. The model is trained from a dataset of short video sequences each capturing a single identity. The training is self-supervised and does not require labeled data. We rely on neural rendering in a one-shot setting without using a 3D graphics model of the human face. 

By using pre-trained components of our model, we avoid the complicated training of a generative model. Our results confirm high flexibility of the StyleGAN2 model, which produces various poses and facial expressions, and that the output can be efficiently controlled by another face of a different identity. 

%By eliminating the need for a 3D graphics model and labeled data, our approach provides a more efficient and practical solution for face synthesis. 
Our main contributions are: (1) Method for pose and expression transfer with close to real-time inference. (2) A Generative model that allows the synthesizing of random identities with controllable pose and expression. 

%The remainder of the paper is structured as follows. We review the existing methods in Sec.~\ref{sec:sota}. The method is presented in Sec.~\ref{sec:method}. Experiments evaluating pose and expression transfer fidelity as well as face identity preservation are given in Sec.~\ref{sec:experiments}. Sec.~\ref{sec:conclusion} concludes the paper. 

%-------------------------------------------------------------------------
\section{Related Work} \label{sec:sota}

Before deep learning methods, the problem of expression transfer was often approached using parametric models. The 3D Morphable Model (3DMM)~\cite{blanz1999morphable} was used in e.g.,~\cite{thies2015real, thies2016face2face}. 

More recently deep models have become prominent. For instance, X2Face \cite{wiles2018x2face} demonstrates that an encoder-decoder architecture with a large collection of video data can be trained to synthesize human faces conditioned by a source frame without any parametric representation of the face or supervision. Furthermore, the paper shows that the expression can be driven not only by the source frame but also by audio to some degree of accuracy.
Similarly,~\cite{zakharov2019few} employs a GAN architecture with an additional embedding network that maps facial images with estimated facial landmarks into an embedding that controls the generator. This allows for conditioning the generated image only on facial landmarks. 

The approach proposed in \cite{wang2021one} enables the generation of a talking-head video from a single input frame and a sequence of 3D keypoints, learned in an unsupervised way, that represent the motions in the video. By utilizing this keypoint representation, the method can efficiently recreate video conference calls. Moreover, the method allows for the extraction of 3D keypoints from a different video, enabling cross-identity motion transfer.

Recently, Megaportraits \cite{drobyshev2022megaportraits} have achieved an impressive level of cross-reenactment quality in one shot.
Their method utilizes an appearance encoder, which encodes the source image into a 4D volumetric tensor and a global latent vector, and a motion encoder, which extracts motion features from both of the input images. These features together with the global latent vector predict two 3D warpings. The first warping removes the source motion from the volumetric features, and the second one imposes the target motion. The features are processed by a 3D generator network and together~with the target motion are input into a 2D convolutional generator that outputs the final image. Their architecture is complex and is made up of many custom modules that are not easily reproducible. Our model is much simpler since it is composed of well-understood open-source publicly available models. We rely on pre-trained StyleGAN2~\cite{karras2020analyzing} to generate the final output and pre-trained ReStyle image encoder~\cite{alaluf2021restyle} to project real input images into the latent space.

Regarding image editing in the latent space of GANs, paper~\cite{radford2015unsupervised} pointed out the arithmetic properties of the generator's latent space. Since then, researchers have extensively studied the editing possibilities that can be done in this domain.
Specifically for StyleGAN, many works have been published regarding latent space exploration \cite{harkonen2020ganspace,shen2020interpreting,abdal2021styleflow,abdal2020image2stylegan++,FaceImageEditing}. InterFaceGAN \cite{shen2020interpreting} shows that linear semantic directions can be easily found in a supervised manner. However, the latent directions are heavily entangled, meaning that one learned latent direction will likely influence other facial attributes as well. 
For example, given a learned latent direction of a pose change, when applied, the person might change expression, hairstyle, or even identity. However, manipulating real input images requires mapping them to the generator's latent space.

%However, if we want to generate and manipulate a particular person, we first have to invert the given image.

The process of finding a latent code that can generate a given image is referred to as the image inversion problem \cite{creswell2018inverting,zhu2016generative,Subrtova22-ECCV}.
% There are two latent spaces considered for this task. The native StyleGAN latent space \( \mathcal{W} \) where a given 512-dimensional latent code \textbf{w} is shared across all of the generator's layers. The other one is the extended latent space where each layer is considered separately, resulting in a larger extended latent space \( \mathcal{W}^+ \) of \( 18\times512 \) dimensions. It has been shown that for the purpose of inverting an image the extended latent space produces better results \cite{abdal2019image2stylegan}.
There are mainly two approaches to image inversion. Either through direct optimization of the latent code to produce the specified image \cite{abdal2020image2stylegan++,abdal2019image2stylegan,roich2022pivotal,zhu2020improved} or through training an encoder on a large collection of images \cite{richardson2021encoding,alaluf2021restyle,tov2021designing}. Typically, direct optimization gives better results, but encoders are much faster. In addition, the encoders show a smoother behavior, producing more coherent results on similar inputs \cite{tzaban2022stitch}.
% It has been demonstrated \cite{zhu2020improved,tov2021designing} that in comparison with \( \mathcal{W^+} \), \( \mathcal{W} \) provides a higher degree of editability, meaning latent codes in this space can be more easily manipulated while maintaining a greater level of realism. However, \( \mathcal{W} \) has poor expressiveness, resulting in inversions that are often inconsistent with target identity. Therefore there exists the so-called distortion-editability trade-off~\cite{zhu2020improved}. Recently paper \cite{roich2022pivotal} has shown  that this trade-off can be bypassed by using PTI (Pivotal Tuning Inversion). The idea is that one may fine-tune the generator around an initial latent code called the pivot. This achieves state-of-the-art inversion and a high level of editability. However, this approach requires storing corresponding generator weights for each individual inversion.

Another reason why we chose to use an encoder for the image inversion is that we require many training images to be inverted and direct optimization of each training sample would not be computationally feasible. We chose ReStyle \cite{alaluf2021restyle}, which uses an iterative encoder to refine the initial estimate of the latent code. This approach is a suitable fit for our purpose, as it leverages smoother behavior over similar inputs from encoders as well as better reconstruction quality from iterative optimization. Currently, the encoders supported in ReStyle are pSp (pixel2style2pixel) \cite{richardson2021encoding} and e4e (encoder4editing) \cite{tov2021designing}. Although both encoders embed images into the extended latent space \( \mathcal{W^+} \), Tov et al. \cite{tov2021designing} argue that by designing an encoder that predicts codes in \( \mathcal{W^+} \) which reside close to \( \mathcal{W} \) they can better balance the distortion-editability trade-off. However, we chose to use ReStyle with a pSp encoder in our network as the baseline method with the e4e encoder had trouble preserving the target identity. 

%Closest to our work is the paper by Yang et al. \cite{yang2019unconstrained} which utilizes StyleGAN for expression transfer. However, their approach heavily relies on direct optimization of the latent code making the method very slow and ineffective for producing videos.
An approach similar in spirit to ours, in the sense of using StyleGAN for expression transfer, is taken by Yang et al.~\cite{yang2019unconstrained}. Nevertheless, they do not transfer the pose, but the expression only. Moreover, their method relies on optimization, which is much slower. They report running times for a single image in minutes, while our method runs in fractions of seconds and is thus more practical for generating videos. 

% --------------------------------------------------------------------------------------------------------
 
\section{Method} \label{sec:method}
\begin{figure*}
\begin{center}
\includegraphics[width=0.7\linewidth]{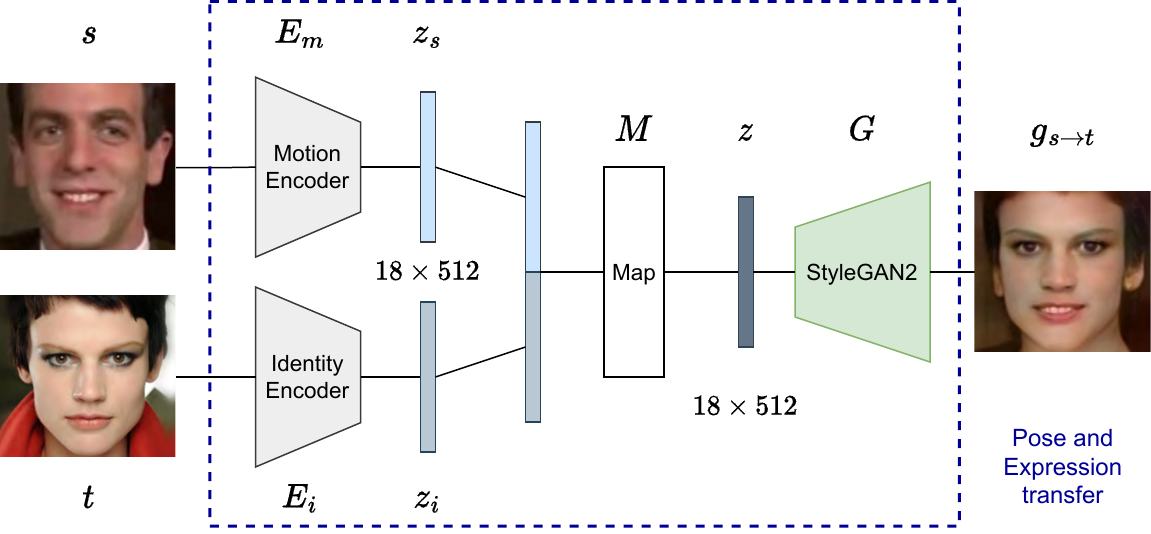}
\end{center}
   \caption{The architecture of the proposed model. The Motion encoder and Mapping network weights are trained, while the Identity encoder and StyleGAN2 weights stay fixed during training.}
\label{fig:architecture}
\end{figure*}
% \begin{figure}[]
%     \centering
%     \includegraphics[width=0.5\textwidth]{iccv2023AuthorKit/Untitled Diagram.drawio (5)1.drawio.pdf}
%     \caption[Network Architecture]{The architecture of the proposed model. The Motion encoder and Mapping network weights are being optimized, while the Identity encoder and StyleGAN2 weights stay fixed during training.}
%     \label{fig:architecture}
% \end{figure}
Our framework takes two face images as input, a source (driving) face image, and a target (identity) face image. The network produces an output image where the pose and expression from the source face image are transferred onto the target identity.
% Our framework takes two input images, a source (driving) face image and a target (identity) face image, and encodes them into latent space embeddings. Both embeddings are then mapped into the latent space of the pre-trained StyleGAN2 \cite{karras2020analyzing}, which generates the output face image.
\subsection{Architecture}
Fig.~\ref{fig:architecture} depicts the proposed architecture. The network consists of a motion (pose+expression) encoder $E_m$, an identity encoder $E_i$, a mapping network $M$, and a generator network $G$.
The encoder $E_i$ embeds the identity of the target face image. The encoder $E_m$ embeds motion, the pose and expression of the source face image. The mapping network then mixes the two embeddings and projects the output into the latent space of the pre-trained StyleGAN2 generator. This approach offers the advantage of generating high-quality images through StyleGAN while avoiding the intricate GAN training process. The network architecture is inspired by \cite{9667038}.
%The advantage of this approach is that StyleGAN generates high-quality images while and the intricate training of GAN is avoided.

Specifically, a source image $s$ and a target image $t$ are aligned and resized to \( 256\times256 \) pixels and then fed into their corresponding encoders, where they are embedded in the extended latent space \( \mathcal{W^+} \) of \( 18\times512 \) dimensions. Embeddings $z_s$ for pose and expression of source image $s$ and $z_t$ for the identity of target image $t$ are then concatenated and transformed through the mapping network into a latent code $z$ \( \in\mathcal{W^+} \) that is then used as an input for the generator that finally produces an output image $g$. Formally, 
\[ g_{s \rightarrow t} = G\bigg(M\Big(E_m(s)\  \oplus \  E_i(t)\Big)\bigg), \] where symbol \(\oplus\) denotes concatenation.

% The architecture utilized for $E_m$ is a ResNet-IR SE 50, which was adapted from the repository published alongside the pSp paper \cite{richardson2021encoding}. It has been shown that this network can be trained to embed various things into the latent space of StyleGAN2 such as cartoons \cite{richardson2021encoding}, hair \cite{9667038} and much more.

ResNet-IR SE 50 has been shown to embed various entities into the latent space of StyleGAN2 such as cartoons \cite{richardson2021encoding}, hair \cite{9667038} and much more. Therefore, we utilize this network as encoder $E_m$. %We adapt it from the repository published along with the pSp paper~\cite{richardson2021encoding}. 
For the encoder $E_i$, we use a pre-trained ReStyle with the pSp configuration. For the mapping network $M$, we employ a single fully connected linear layer. For the generator, we use the pre-trained StyleGAN2 which produces high-resolution images of \( 1024\times1024 \) px.
%-------------------------------------------------------------------------
\subsection{Training}
We employ self-supervised training to optimize the parameters of the encoder $E_m$ and the mapping network $M$, while keeping the parameters of the generator $G$ and the encoder $E_i$ fixed. The training is performed on an unlabeled dataset of short video clips, each containing a single person.

% We optimize the parameters of the encoder $E_m$ and the mapping network $M$. The paramters of the generator $G$ and the encoder $E_i$  stay fixed. The training is self-supervised. 
% An unlabaled dataset of short video clips is used where in each video clip only one person is present.
During each iteration of the training procedure, we randomly sample two pairs of frames ($s_A$, $t_A$) and ($s_B$, $t_B$) from two video clips of identities $A$ and $B$, respectively.  %More specifically, one source frame $s_A$ and one target frame $t_A$ of identity A and a source frame $s_R$ and a target frame $t_R$ of a random identity.
%More specifically, a source and target frame pair ($s_A$, $t_A$) of identity A are randomly sampled from a video clip, and another source and target frame pair ($s_R$, $t_R$) of a random identity are randomly sampled from a different video. 
We then generate two images $g_{s_{A}\rightarrow t_{A}}$ where the source and target frames are of identity $A$ and $g_{s_{A}\rightarrow t_{B}}$ where the source is of identity $A$ and the target is of identity $B$. %The notation $g_{s_{A}\rightarrow t_{R}}$ implies that the pose and expression from the source image $s_A$ is imposed onto the indentity R from the target image $t_R$.
We employ the following loss functions:
\begin{itemize}[label={},leftmargin=0cm]
\item \textbf{Pixel-wise loss}.\, It is Euclidean distance between the source and generated image intensities
\begin{equation}
     \mathcal{L}_{2} =  \| s_A-g_{s_{A}\rightarrow t_{A}} \|_{2}. 
\end{equation}
   where $s_A$ is the source frame of identity $A$ and $g_{s_{A}\rightarrow t_{A}}$ is a generated image using both inputs from identity $A$.
\item \textbf{Perceptual loss}.\, LPIPS (Learned Perceptual Image Patch Similarity)~\cite{lpips} was shown to correlate with human perception of image similarity. In praticular,  %The pixel-wise loss falls short in capturing the perceptual changes that humans notice when viewing images.
%To address this limitation, perceptual loss has been developed. Instead of analyzing individual pixels, perceptual loss compares the higher-level similarities between two images, such as their content and style.
%LPIPS (Learned Perceptual Image Patch Similarity) loss has been proposed in \cite{lpips}.
   % This can be demonstrated on blurring an image, which may not cause a significant change in the \(\mathcal{L}_2\) value but can still be visually noticeable to a human eye. To address this limitation, the perceptual loss has been developed. Instead of analyzing individual pixels, perceptual loss compares the higher-level similarities between two images, such as their content and style. LPIPS (Learned Perceptual Image Patch Similarity) loss has been proposed in \cite{lpips}, which involves utilizing a dataset of almost half a million human judgments to compute the perceptual distance between reference and generated images. This approach improves upon the limitations of the pixel-wise function and better captures the perceptual differences that humans notice when viewing images. We employ a pre-trained neural network to calculate the perceptual distance 
  \begin{equation}
   \mathcal{L}_{LPIPS} =  1 - \langle P(s_A), P(g_{s_{A}\rightarrow t_{A}}) \rangle, \end{equation} 
  where $P$ is a perceptual feature extractor (AlexNet)~\cite{NIPS2012_c399862d} that outputs unit-length normalized features and \(\langle.,. \rangle\) denotes dot product. % We again use only images of the same identity in this loss.
  %and therefore the entire loss function can be described as a cosine distance of the features.
  \item \textbf{Identity loss}. To ensure that the generated image preserves the identity of the target image, we employ the pre-trained facial recognition model ArcFace~\cite{arcface}. We calculate it in a similar fashion to the previous loss:
  \begin{equation}
   \mathcal{L}_{ID} =  1 - \langle D(t_B),D(g_{s_{A}\rightarrow t_{B}})\rangle , \end{equation}
  where $D$ produces unit-length normalized embeddings of respective frames. %, $g_{s_A}$ is the inverted source frame of identity A (Image generated with the latent code produced by $E_i$ from $s_A$), and $g_{s_{A}\rightarrow t_{R}}$ is a generated image where the source frame is of identity A and the target frame is of identity R. 
\item \textbf{CosFace loss}.\, Finally, we implement the CosFace loss~\cite{cosface} that we use in a similar way to Megaportraits~\cite{drobyshev2022megaportraits}. The purpose of the loss is to make the embeddings of coherent pose and expressions similar, while maintaining the embeddings of independent pose and expressions uncorrelated. 
%
%The idea behind this loss is that when we randomly sample a source frame from each of the two different videos, the pose and expression in these frames will likely differ. In that case, the generated image, using one source frame, should have a similar motion descriptor to that particular source frame, while also having a dissimilar motion descriptor to the other source frame regardless of the identity used to generate the image.
%
For this loss, only motion descriptors embedded by $E_m$, are necessary. We calculate motion descriptors $z_A=E_m(s_{A})$, $z_B=E_m(s_{B})$ of the inputs, and of the outputs fed to the encoder  $z_{A\rightarrow A}=E_m(g_{s_{A}\rightarrow t_{A}})$,  $z_{A\rightarrow B}=E_m(g_{s_{A}\rightarrow t_{B}})$. We then arrange them into positive pairs $P$ that should align with each other: $P = {(z_{A\rightarrow A},z_A),(z_{A\rightarrow B},z_A)}$, and negative pairs: $N = {(z_{A\rightarrow A},z_B),(z_{A\rightarrow B},z_B)}$. These pairs are then used to calculate the following cosine distance: \begin{equation} d(z_i,z_j) =  a \cdot(\langle z_i,z_j \rangle - b), \end{equation} where both $a$ and $b$ are hyperparameters. Finally, % This distance is then used to calculate the Large Margin Cosine Loss 
%For this loss, only motion descriptors which are embedded by $E_m$, are necessary. We separately calculate a motion descriptor $z_B=E_m(s_{B})$ while also storing the motion descriptor calculated during the forward pass of the network, $z_A=E_m(s_{A})$. Lastly, the motion descriptors of the generated images $z_{A\rightarrow A}=E_m(g_{s_{A}\rightarrow t_{A}})$ and $z_{A\rightarrow B}=E_m(g_{s_{A}\rightarrow t_{B}})$ are calculated. We then arrange them into positive pairs $P$ that should align with each other: $P = {(z_{A\rightarrow A},z_A),(z_{A\rightarrow B},z_A)}$, and negative pairs: $N = {(z_{A\rightarrow A},z_B),(z_{A\rightarrow B},z_B)}$. These pairs are then used to calculate the following cosine distance: \begin{equation} d(z_i,z_j) =  a \cdot(\langle z_i,z_j \rangle - b), \end{equation} where both $a$ and $b$ are hyperparameters. Finally, % This distance is then used to calculate the Large Margin Cosine Loss \cite{cosface}:
%
  \begin{equation}
  \mathcal{L}_{cos} \!=\! - \!\!\!\!\!\!\! \sum_{(z_k,z_l)\in\mathcal{P}} \!\!\!\!\!\!\!{\log{\frac{\mathrm{exp}\{d(z_k,z_l)\}}{\mathrm{exp}\{d(z_k,z_l)\}+\!\!\!\!\!\!\!\!\!\sum\limits_{(z_i,z_j)\in\mathcal{N}}\!\!\!\!\!\!\!{\mathrm{exp}\{d(z_i,z_j)\}}
}}}.
  \end{equation}

\end{itemize}
Furthermore, we used cropped versions of the $\mathcal{L}_{2}$ loss and the $\mathcal{L}_{LPIPS}$ losses. The crop is the central area of \( 188\times188 \) pixels of the original \( 256\times256 \) aligned image. The losses $\mathcal{L}_{2\_crop}$ and $\mathcal{L}_{LPIPS\_crop}$ are used exactly as their aforementioned counterparts. The cropped losses turned out to be important. Otherwise, we observed the model struggled to transfer the expression precisely, probably being disturbed by the complex texture of hair and background. 

The total loss which is used to train the network is the weighted sum of the individual losses
\begin{equation}
\begin{split}
    \mathcal{L} = w_{\mathcal{L}_2}\mathcal{L}_2 + w_{LPIPS}\mathcal{L}_{LPIPS} + w_{ID}\mathcal{L}_{ID} \\ + w_{cos}\mathcal{L}_{cos} + w_{\mathcal{L}_{2\,\_crop}}\mathcal{L}_{2\,\_crop} \\ + w_{LPIPS\_crop}\mathcal{L}_{LPIPS\_crop}.
\end{split}
\end{equation}

\subsection{Dataset}\label{sec:dataset}
For our goal, we need a dataset consisting of numerous unique identities and a wide range of images with varying poses and facial expressions for each identity. To meet this requirement, it was necessary to use video data despite a potential trade-off in image quality. 

We decided to use the VoxCeleb2 dataset \cite{voxceleb2} which was collected originally for speaker recognition and verification. It has since been used for talking head synthesis, speech separation, and face generation. It contains over a million utterances from 6 112 identities, providing us with a vast array of subjects to work with. The dataset is primarily composed of celebrity interview videos, offering a broad spectrum of poses and expressions to utilize. The videos are categorized by identity and trimmed into shorter utterances that range from 5 to 15 seconds in duration. They have also already undergone preprocessing that includes cropping the frames to the bounding boxes around each speaker's face.  On top of that, we use the official preprocessing script provided by StyleGAN to normalize the images to $224\times224$ pixels~\cite{ffhq}. 

%Unfortunately, the preprocessing step does not match the one required by StyleGAN as the faces are cropped by the forehead. StyleGAN requires the images to be of the entihead, head including the top part that is missing. We use the official preprocessing script provided by StyleGAN that pads the missing part of the forehead. This results in all of our training and testing images having blurred stripes at the top. Although this hinders the quality of the inverted images as well as the generated ones, it is better than the alternative datasets that we considered. The alternatives did not have a large collection of speakers and they were not segregated based on identity, resulting in multiple individuals speaking in a single utterance.

%Another challenge with the dataset is the relatively low resolution of the videos, typically $224\times224$ pixels. This is problematic since StyleGAN is designed to generate high-quality images with a resolution of $1024\times1024$ pixels. Nevertheless, finding a large dataset of high-resolution videos featuring a vast number of distinct individuals is nearly impossible.

As the number of videos per individual differs, we balanced it out by only using a maximum number of videos per person. We extracted 10 frames at half-second intervals from each video. Subsequently, we eliminate images with extreme poses that would be difficult to generate with StyleGAN. %We also pre-align all of the images using the official StyleGAN preprocessing script, which uses dlib \cite{dlib09}, a machine-learning library, to detect human faces and Facial Landmarks. If the image has such a bad quality that the face or the Facial Landmarks are not detected, the image is dropped as well. 
The final training set contains around 6k different identities, each with around 10 images from 5 different video clips, resulting in a little under 300k images. The dataset was split into disjoint training-validation-test sets 80-10-10 percent, respectively. No identity appears in any of the splits simultaneously.
%Our filtered training dataset contains a little under 300 000 images of 6 000 different identities. 

\subsection{Implementation details}\label{sec:train_details}
%The model, as well as the data set, is very large, so we trained it for a million steps with a batch size of 8. Surprisingly, even after that many steps, the validation loss kept slowly decreasing. However, we observed a decline in the ability of the model to accurately capture facial expressions in the generated images around the one million training step mark, despite the decreasing validation loss. 
%The loss function does not capture the expression perfectly and that is the reason why we had to revisit all of the intermediate model weights and check if there are better ones even with higher validation loss. We chose the model weights which transferred the expression the best on the validation set.
%ikdyz klesa ztratova funkce tak jsme pozorovali ze kolem 1 mil trenovacich kroku ze se zhorsovat prenos vyrazu ikdyz klesa ztrat. funkce, protoze nezachycuje presne vyraz a proto to vybereme podle nediferencovatelnych metrik na valdiacnich datech.
% We attempted to employ a loss function to specifically model facial expressions, but it resulted in an even worse performance, which will be discussed in the next chapter. Because of this, we had to revisit all of the intermediate model weights and check if there are better ones even with higher validation loss. We chose the model weights which transferred the expression the best based on the evaluations discussed in \ref{sec:eval-quantitative}.
% % training steps  misto iteration

The model was trained for about a million steps with a batch size of 8. The best model checkpoint was selected based on the error statistics measuring the expression transfer fidelity and identity preservation, see Sec.~\ref{sec:eval-quantitative}. 

We used the ranger optimizer \cite{ranger}, which combines the Rectified Adam algorithm and Look Ahead. We set the learning rate to $1\cdot10^{-5}$. For our model with the best performance, we used the following hyperparameters for the losses: \( w_{\mathcal{L}_2} = 0\), \( w_{LPIPS} = 0.05\), \( w_{ID} = 0.3\), \( w_{cos} = 0\), \( w_{\mathcal{L}_{2\,\_crop}} = 2\), \( w_{LPIPS\_crop} = 0.3\). We set parameters $a = 5$ and $b = 0.2$ in the CosFace loss. 
%The parameters of our best model heavily rely on crop versions of the self-reenactment losses. The reason for this is that when we tried to use full images as input for those losses, the network struggled to learn the desired facial expression manipulation. It instead had to focus on the background and hair fidelity and thus failed to transfer the expressions correctly. 

\section{Experiments} \label{sec:experiments}

\subsection{Comparison of methods}\label{sec:eval-method}

\paragraph{Baseline method.}
To the best of our knowledge, we are not aware of any publicly available implementation of our problem. Therefore, we compare the proposed method with a linear StyleGAN latent space manipulation as the baseline method. 

%For our baseline method, we take advantage of the arithmetic property of the StyleGANs latent space. As mentioned before, the latent space has a linear property, where the latent codes can be added and subtracted for meaningful edits. However, these edits have difficulties preserving the identity of the person.% and in extreme cases, a person can grow a beard, change hairstyle, change the facial structure, or even produce undesirable artifacts that disrupt the image. 

Given two frames $A_0$ and $A_1$ (sampled from the same video) where the pose and expression of the person differ, the edit vector is represented by the difference between the latent codes corresponding to the inverted frames. The pose and expression can then be imposed on a different person in image $B$ by adding the edit vector to the latent code of image $B$. Formally, 
\begin{equation} z_{A_{1}\rightarrow B} = z_B + \alpha \cdot (z_{A_{1}}-z_{A_{0}}), \end{equation} 
%$z_{A_{1}\rightarrow B} = z_B + \alpha \cdot (z_{A_{1}}-z_{A_{0}}),$
where $z_B$ is the latent code of the target person,  $z_{A_{0}}$ is the latent code of the person $A$ with the initial pose and expression and $z_{A_{1}}$ is the latent code of the same person with a different pose and expression. Scalar $\alpha$ represents the magnitude of the edit and the resulting latent code $z_{A_{1}\rightarrow B}$ fed into StyleGAN generates the output, ideally a person $B$ with the pose and expression of $A_1$. In our case, we always set $\alpha$ to one, to get the same expression and pose.

However, this approach requires the initial pose and facial expression in frame $A_0$ to match the pose and expression of the person in frame $B$. This is a very strict requirement, as there will probably be no frame in a video where the pose and expression match perfectly. %the pose and expression in frame $B$. %Even when two short videos are considered, there might not be two frames, each from a different video, where the pose and expression match precisely.

Instead of searching for two frames that match pose and expression the best, we utilize an arithmetic property of the latent space. We flip each frame in a video by the vertical axis and invert them along with their non-flipped counterparts. Then we calculate the mean latent code for all the frames. This results in a frontal pose with an average expression across the video, typically a neutral expression. We do this for both videos, which provides us with the same pose and a similar expression for the initial frames. We then used the aforementioned method to transfer pose and expression from one person to another. The downside of this method is that it does not work with single images, but requires a short video of each individual. Moreover, inverting all the frames within the videos is required, which is computationally demanding.

We consider two versions of the baseline method. Both invert all the images with ReStyle \cite{alaluf2021restyle}, but one with the pSp encoder configuration \cite{richardson2021encoding} and the other with the e4e configuration \cite{tov2021designing}.

\paragraph{Variants of our method.}
Besides the default model presented in Sec.~\ref{sec:method} denoted as (Ours), we tested the other two variants. (Ours-Gen) does not have the StyleGAN generator fixed, but its weights are optimized during the training of the entire model. (Ours-Cos) is the model where the CosFace loss is engaged during training. CosFace loss has zero weight and the SyleGAN generator is fixed in the default model.

\subsection{Qualitative evaluation}

\begin{figure*}[ht]
    \centering
    \setlength{\tabcolsep}{0.005\linewidth}
    \renewcommand{\arraystretch}{0.03}
    \begin{tabular}{c@{\hspace{0.03\linewidth}}ccccc}
        \hfill\raisebox{0.7cm}{Target} & 
        \includegraphics[width=0.15\linewidth]{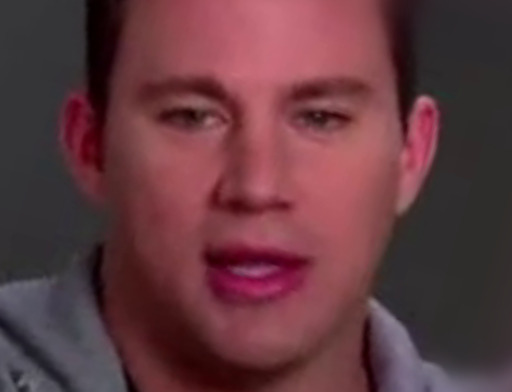} & 
        \includegraphics[width=0.15\linewidth]{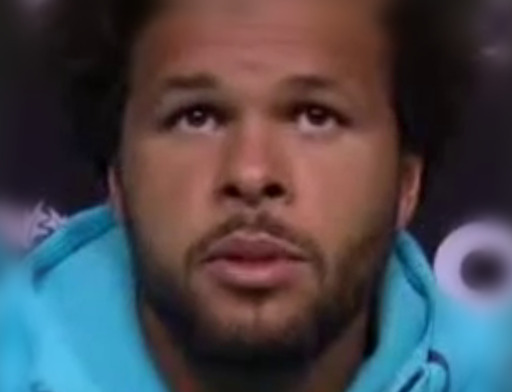} & 
        \includegraphics[width=0.15\linewidth]{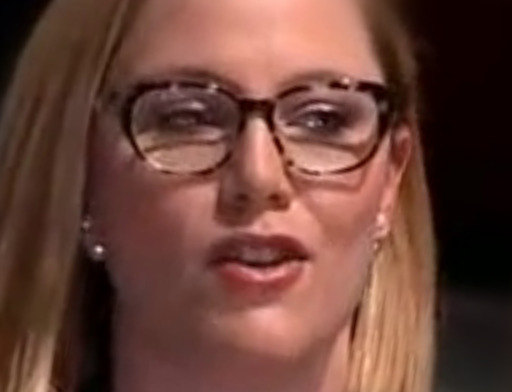} & 
        \includegraphics[width=0.15\linewidth]{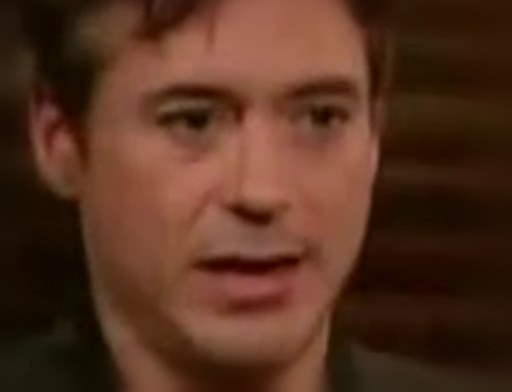} & 
        \includegraphics[width=0.15\linewidth]{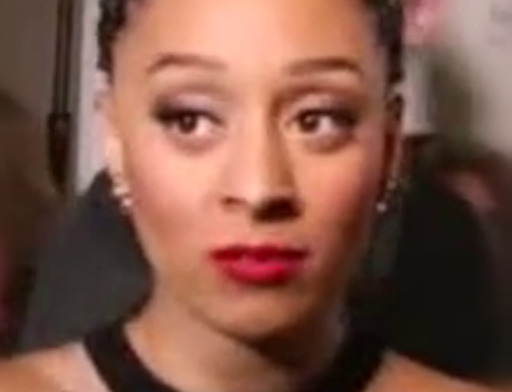} \\
        \vspace{0.03\textwidth} \\
        % \hfill\raisebox{0.7cm}{Inversion} & 
        % \includegraphics[width=0.15\linewidth]{src/images/grid_images/frame88_01latent_crop.png} & 
        % \includegraphics[width=0.15\linewidth]{src/images/grid_images/frame68_01latent_crop.png} & 
        % \includegraphics[width=0.15\linewidth]{src/images/grid_images/frame71_01latent_crop.png} & 
        % \includegraphics[width=0.15\linewidth]{src/images/grid_images/frame1_01latent_crop.png} & 
        % \includegraphics[width=0.15\linewidth]{src/images/grid_images/frame47_01latent_crop.png} \\
        % \vspace{0.03\textwidth} \\
        \includegraphics[width=0.15\linewidth]{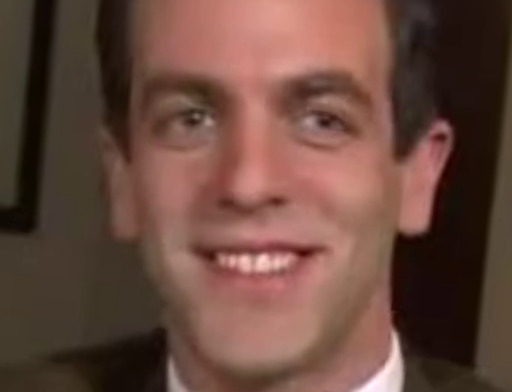} & 
        \includegraphics[width=0.15\linewidth]{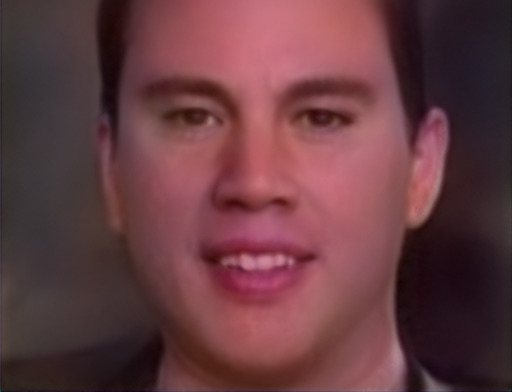} & 
        \includegraphics[width=0.15\linewidth]{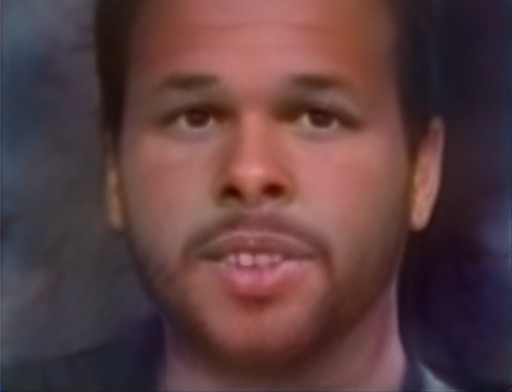} & 
        \includegraphics[width=0.15\linewidth]{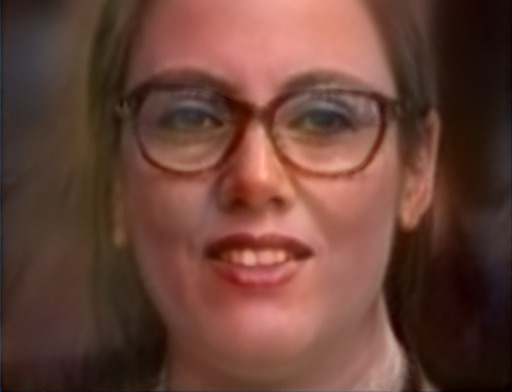} & 
        \includegraphics[width=0.15\linewidth]{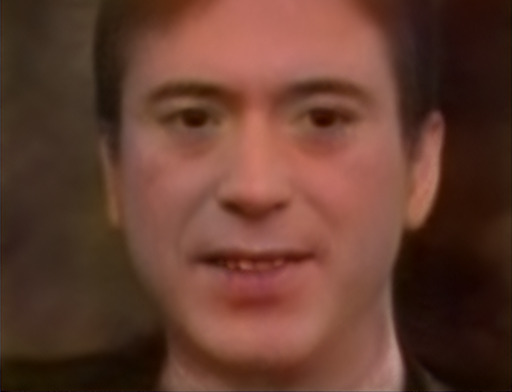} & 
        \includegraphics[width=0.15\linewidth]{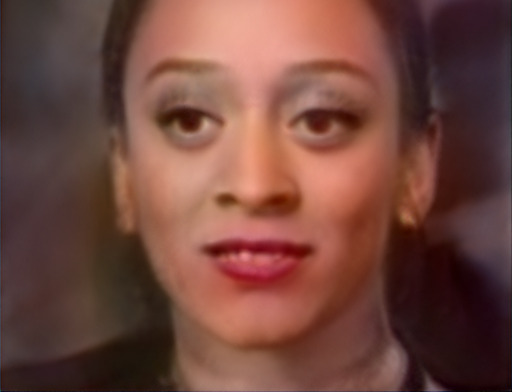} \\
        \vspace{0.005\linewidth} \\
        \includegraphics[width=0.15\linewidth]{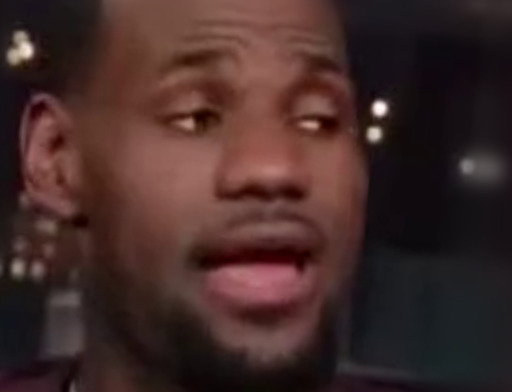} & 
        \includegraphics[width=0.15\linewidth]{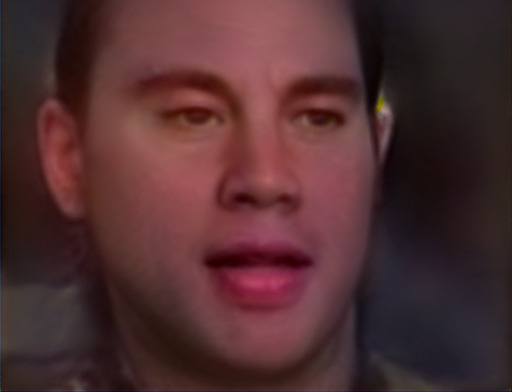} & 
        \includegraphics[width=0.15\linewidth]{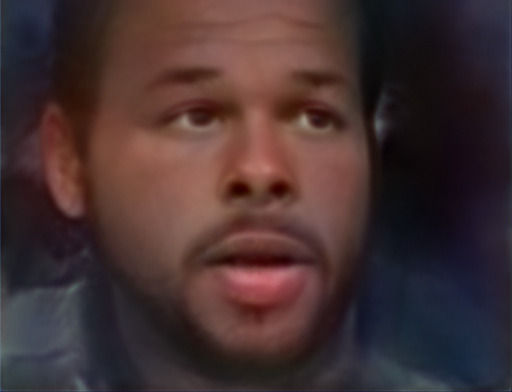} & 
        \includegraphics[width=0.15\linewidth]{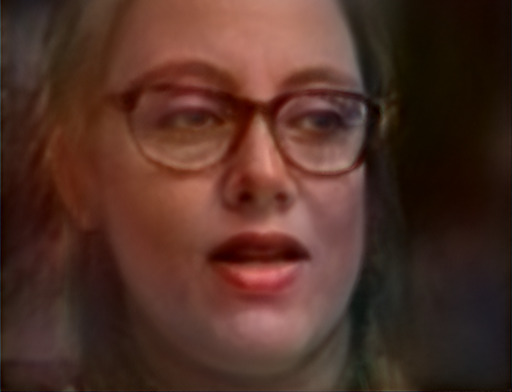} & 
        \includegraphics[width=0.15\linewidth]{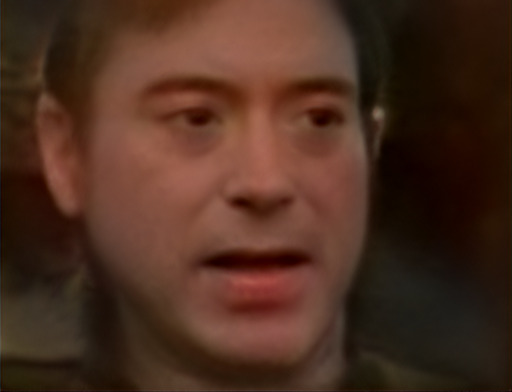} & 
        \includegraphics[width=0.15\linewidth]{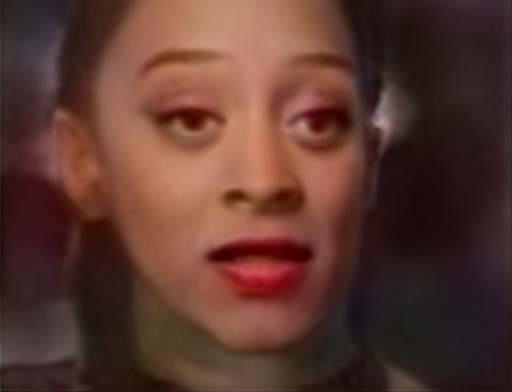} \\
        % \vspace{0.005\linewidth} \\
        % \includegraphics[width=0.15\linewidth]{src/images/grid_images/frame147_01source_crop.jpg} & 
        % \includegraphics[width=0.15\linewidth]{src/images/grid_images/frame147_01-frame88_01.jpg} & 
        % \includegraphics[width=0.15\linewidth]{src/images/grid_images/frame147_01-frame68_01.jpg} & 
        % \includegraphics[width=0.15\linewidth]{src/images/grid_images/frame147_01-frame71_01.jpg} & 
        % \includegraphics[width=0.15\linewidth]{src/images/grid_images/frame147_01-frame1_01.jpg} & 
        % \includegraphics[width=0.15\linewidth]{src/images/grid_images/frame147_01-frame47_01.jpg} \\
        \vspace{0.005\linewidth} \\
        \includegraphics[width=0.15\linewidth]{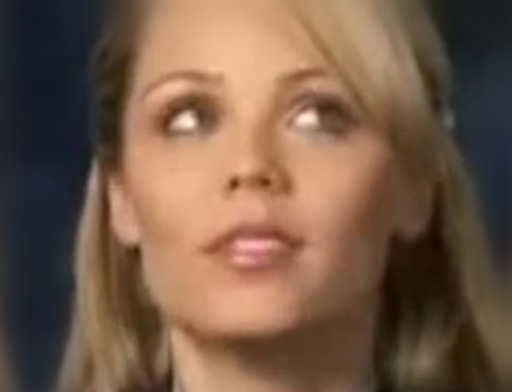} & 
        \includegraphics[width=0.15\linewidth]{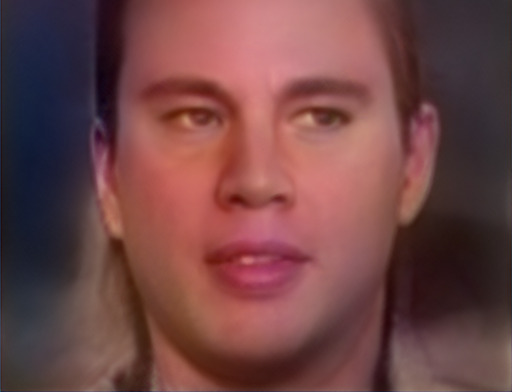} & 
        \includegraphics[width=0.15\linewidth]{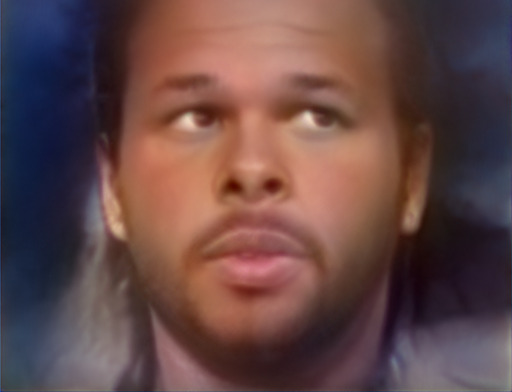} & 
        \includegraphics[width=0.15\linewidth]{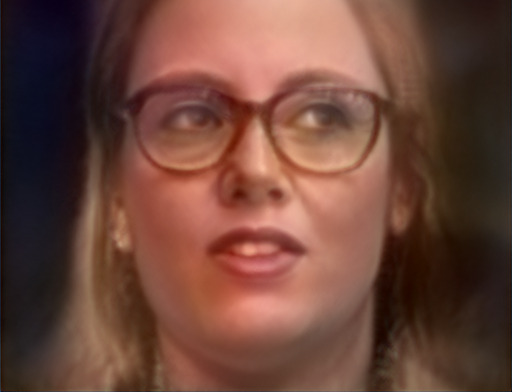} & 
        \includegraphics[width=0.15\linewidth]{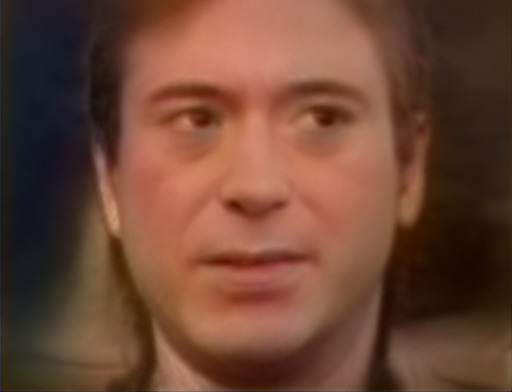} & 
        \includegraphics[width=0.15\linewidth]{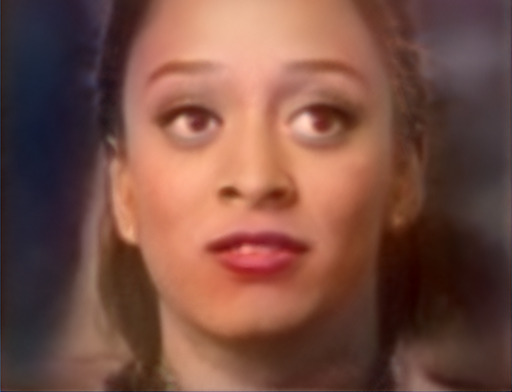} \\
        \vspace{0.005\linewidth} \\
        \includegraphics[width=0.15\linewidth]{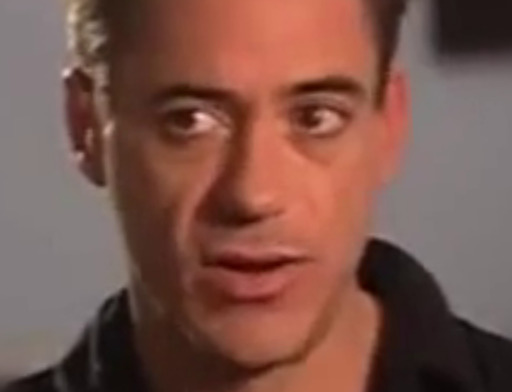} & 
        \includegraphics[width=0.15\linewidth]{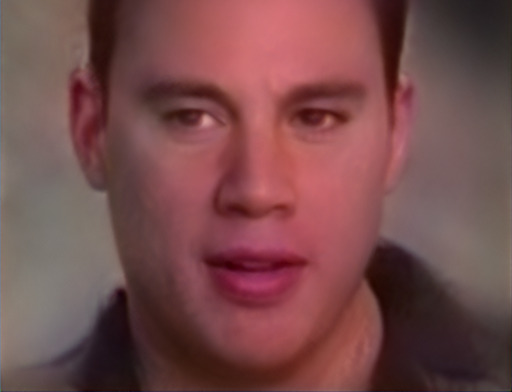} & 
        \includegraphics[width=0.15\linewidth]{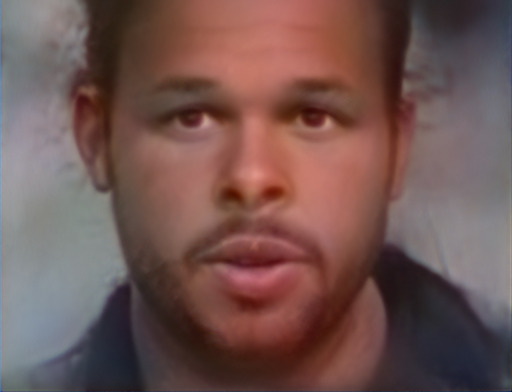} & 
        \includegraphics[width=0.15\linewidth]{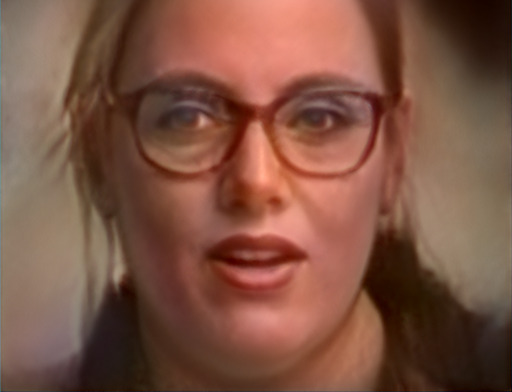} & 
        \includegraphics[width=0.15\linewidth]{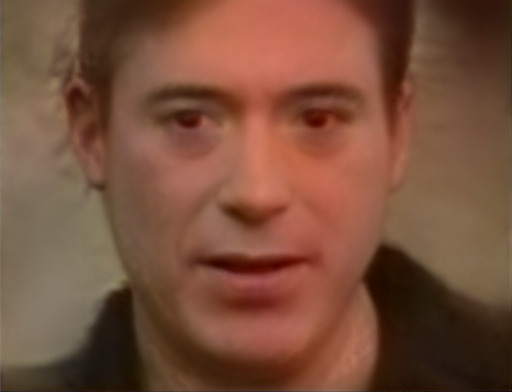} & 
        \includegraphics[width=0.15\linewidth]{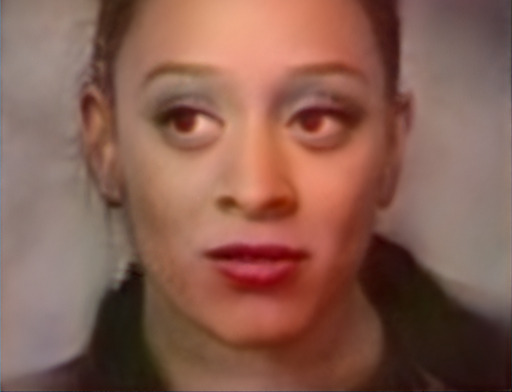} \\
        % \vspace{0.005\linewidth} \\
        % \includegraphics[width=0.15\linewidth]{src/images/grid_images/frame103_01source_crop.png} & 
        % \includegraphics[width=0.15\linewidth]{src/images/grid_images/frame103_01-frame88_01.png} & 
        % \includegraphics[width=0.15\linewidth]{src/images/grid_images/frame103_01-frame68_01.png} & 
        % \includegraphics[width=0.15\linewidth]{src/images/grid_images/frame103_01-frame71_01.png} & 
        % \includegraphics[width=0.15\linewidth]{src/images/grid_images/frame103_01-frame1_01.png} & 
        % \includegraphics[width=0.15\linewidth]{src/images/grid_images/frame103_01-frame47_01.png} \\
        \vspace{0.01\textwidth} \\
        \raisebox{0.9cm}{Source} & 
        \raisebox{0.9cm}{Identity 1} & 
        \raisebox{0.9cm}{Identity 2} & 
        \raisebox{0.9cm}{Identity 3} & 
        \raisebox{0.9cm}{Identity 4} & 
        \raisebox{0.9cm}{Identity 5} \\
    \end{tabular}
    \vspace*{-3mm}
    \caption{Pose and expression transfer results. The top row depicts the target (identity) input images, leftmost column the source (driving) input images. The grid shows the transfer results. The identities are preserved column-wise, and the poses and expressions are preserved row-wise.}
    \vspace*{-3mm}
    \label{fig:tabular_images}
\end{figure*}

In Fig.~\ref{fig:tabular_images} we present several examples of pose and expression transfer between a variety of identities. The pairs are challenging since the input frames differ in ethnicity, gender, and illumination. Another challenge is the accessories that people wear such as glasses or earrings. 

The pose and expression are seen to be transferred while still preserving the input identity. The model learned to transfer pose, expression, and eye movement. The network also correctly identifies that if eyeglasses are present in the identity image, they are preserved in the output image. Surprisingly, the network is able to model eye movement even behind glasses. However, the model is not perfect for preserving hair or background. % correctly as discussed in \ref{sec:train_details}.
%The network does not model the hair or background correctly as discussed in \ref{sec:train_details}.% TODO zkontrolovat toto po predelani experiments
\def \x {0.13}
\begin{figure*}[ht]
    \centering
    \setlength{\tabcolsep}{0.005\linewidth}
    \renewcommand{\arraystretch}{0.03}
    \begin{tabular}{c@{\hspace{0.03\linewidth}}ccccc}
        \hfill\raisebox{0.7cm}{Source} & 
        \includegraphics[width=\x\linewidth]{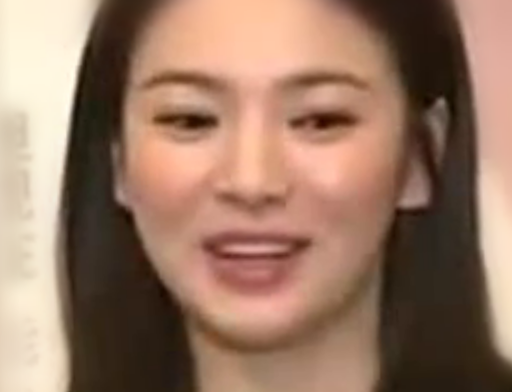} & 
        \includegraphics[width=\x\linewidth]{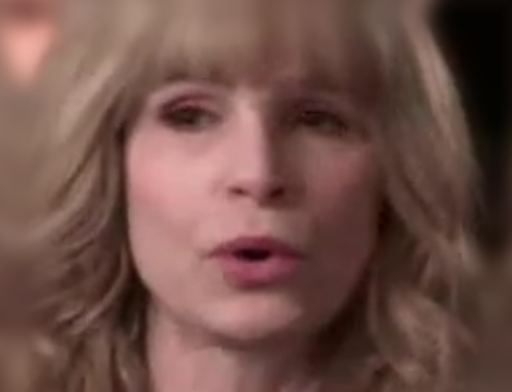} & 
        \includegraphics[width=\x\linewidth]{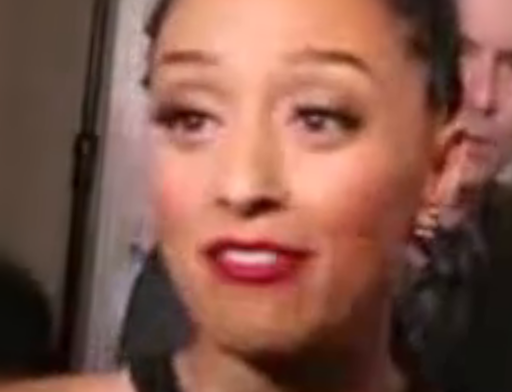} &
        \includegraphics[width=\x\linewidth]{src/images/grid_images/frame40_01source_crop.jpg}  & 
        \includegraphics[width=\x\linewidth]{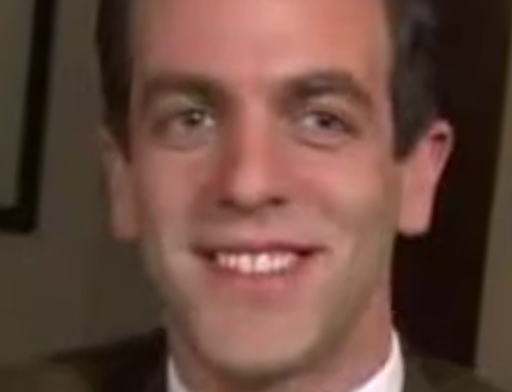}
        \\
        % \vspace{0.03\textwidth} \\
        \hfill\raisebox{0.7cm}{Target} & 
        \includegraphics[width=\x\linewidth]{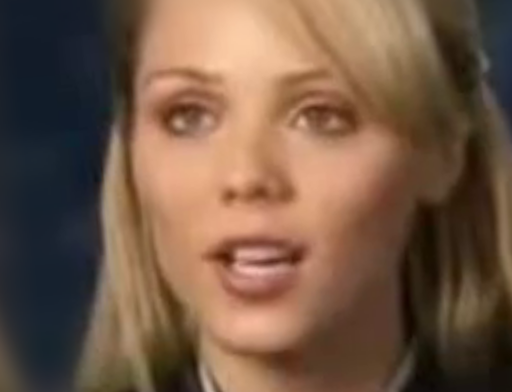} & 
        \includegraphics[width=\x\linewidth]{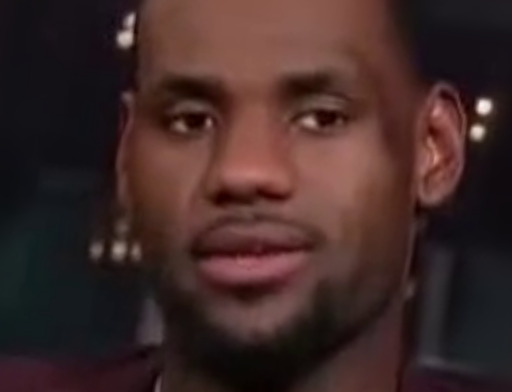} & 
        \includegraphics[width=\x\linewidth]{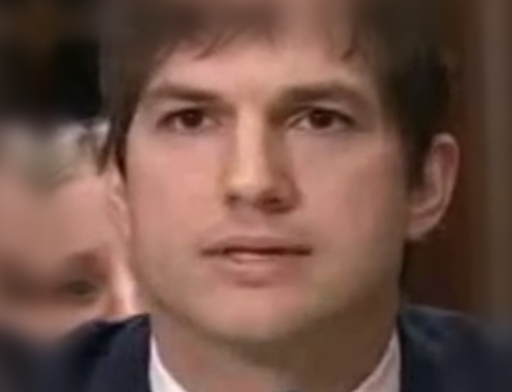} &
        \includegraphics[width=\x\linewidth]{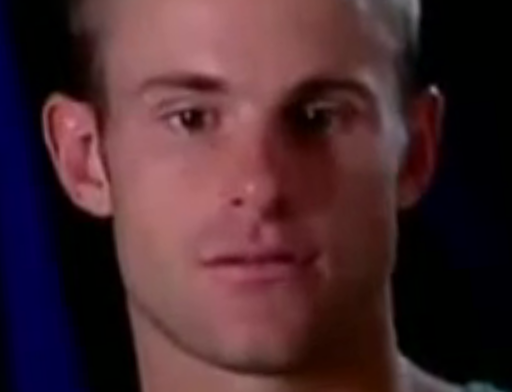}  & 
        \includegraphics[width=\x\linewidth]{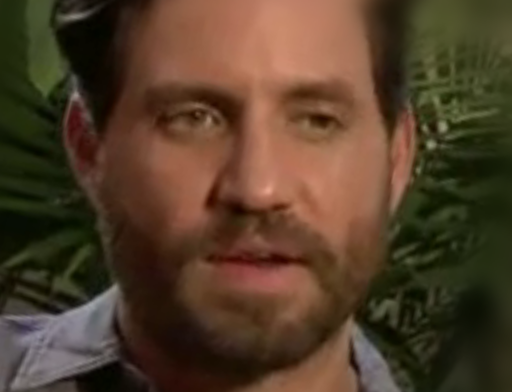}
        \\
        \vspace{0.01\textwidth} \\
        \raisebox{0.7cm}{Base pSp} & 
        \includegraphics[width=\x\linewidth]{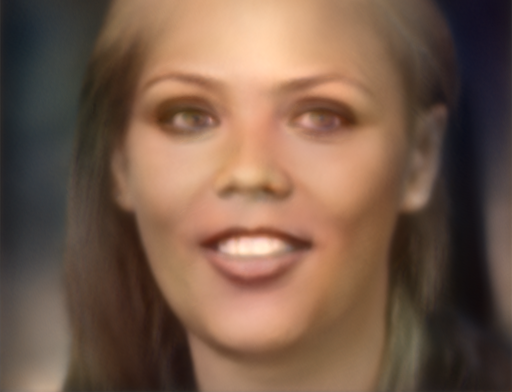}& 
        \includegraphics[width=\x\linewidth]{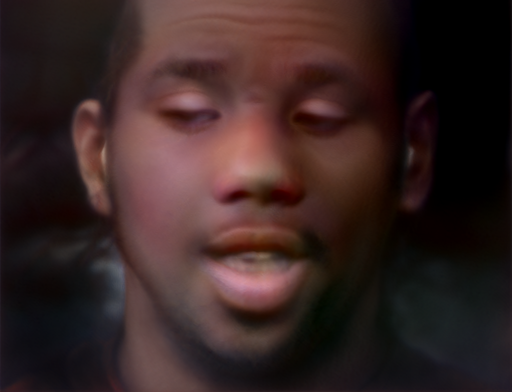} & 
        \includegraphics[width=\x\linewidth]{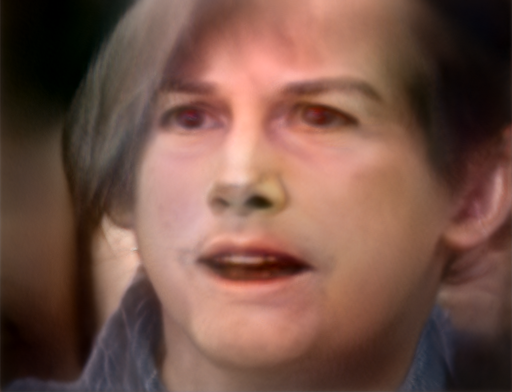} & 
        \includegraphics[width=\x\linewidth]{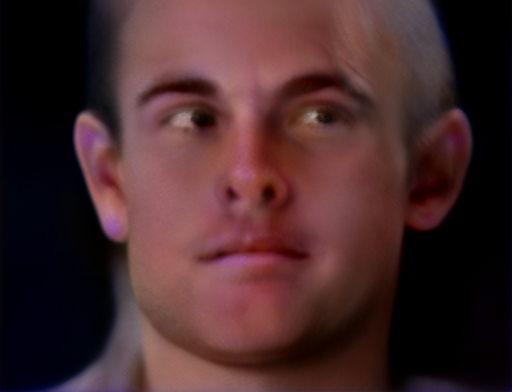} & 
        \includegraphics[width=\x\linewidth]{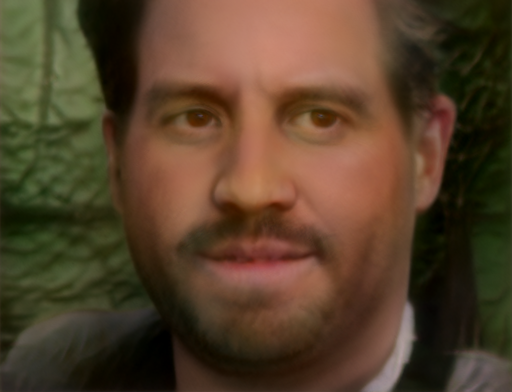} 
        \\
        %\vspace{0.005\linewidth} \\
        \raisebox{0.7cm}{Base e4e} & 
        \includegraphics[width=\x\linewidth]{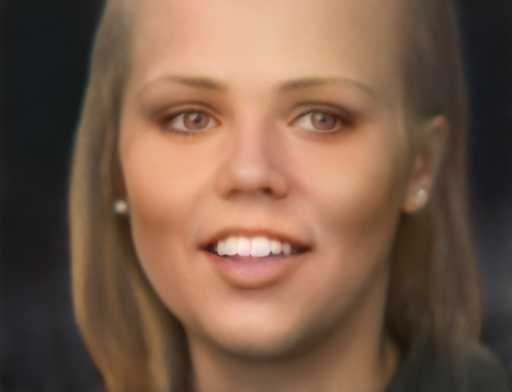}  &  
        \includegraphics[width=\x\linewidth]{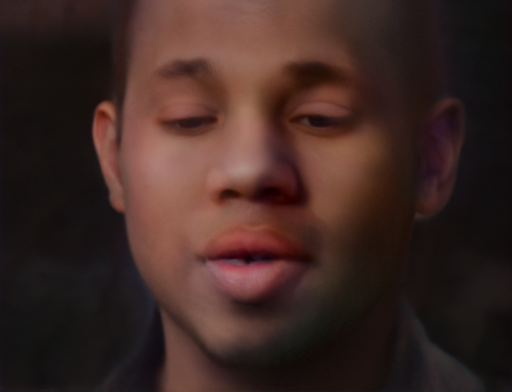} & 
        \includegraphics[width=\x\linewidth]{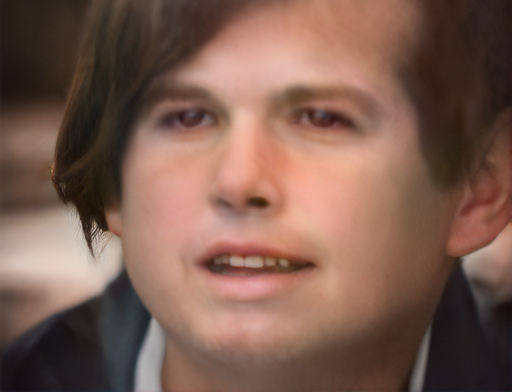} &
        \includegraphics[width=\x\linewidth]{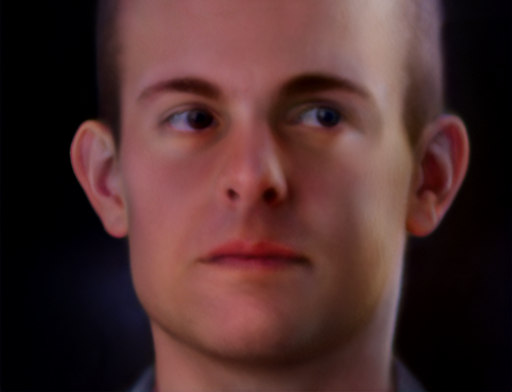} & 
        \includegraphics[width=\x\linewidth]{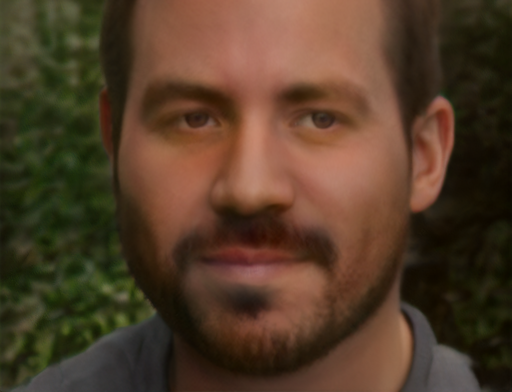}
        \\
        %\vspace{0.005\linewidth} \\
        \raisebox{0.7cm}{Ours Gen} & 
        \includegraphics[width=\x\linewidth]{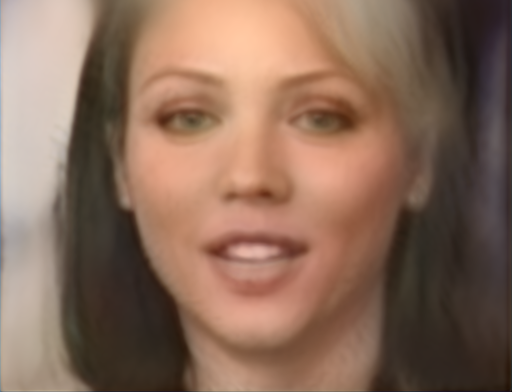} & 
        \includegraphics[width=\x\linewidth]{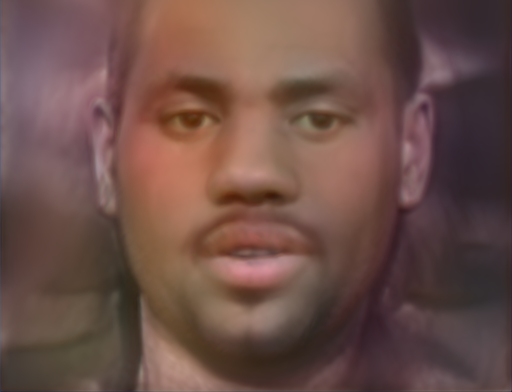} & 
        \includegraphics[width=\x\linewidth]{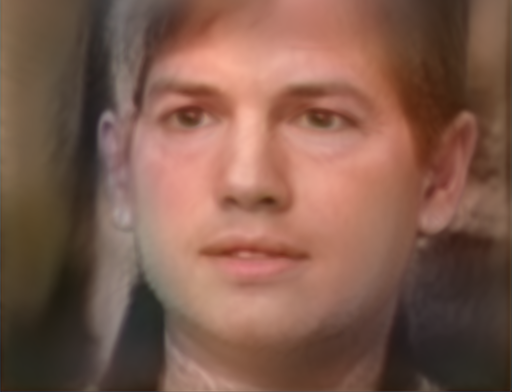} &
        \includegraphics[width=\x\linewidth]{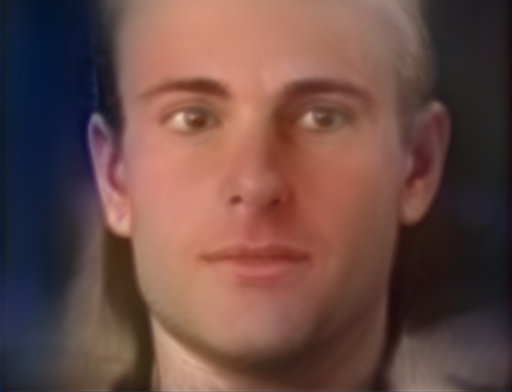}  & 
        \includegraphics[width=\x\linewidth]{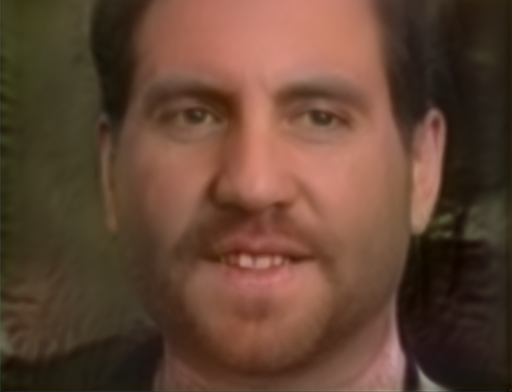}
        \\
        %\vspace{0.005\linewidth}
        \\
        \raisebox{0.7cm}{Ours Cos} & 
        \includegraphics[width=\x\linewidth]{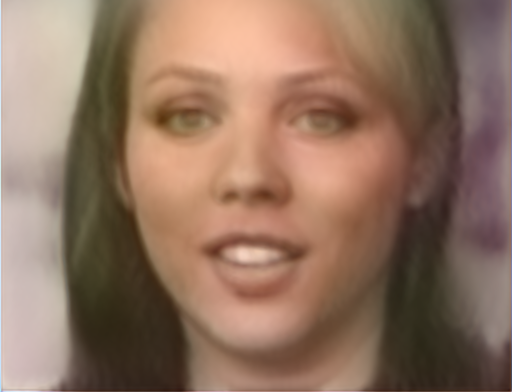} & 
        \includegraphics[width=\x\linewidth]{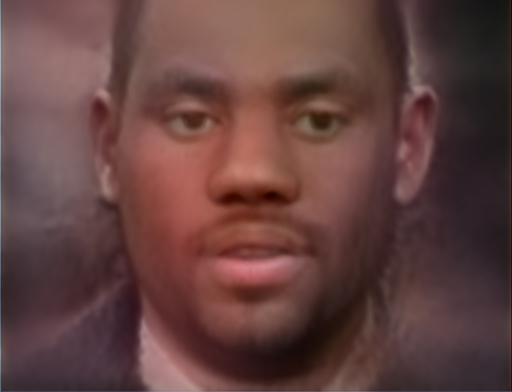}& 
        \includegraphics[width=\x\linewidth]{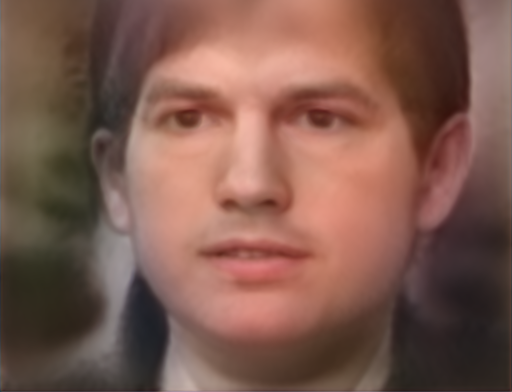} & 
        \includegraphics[width=\x\linewidth]{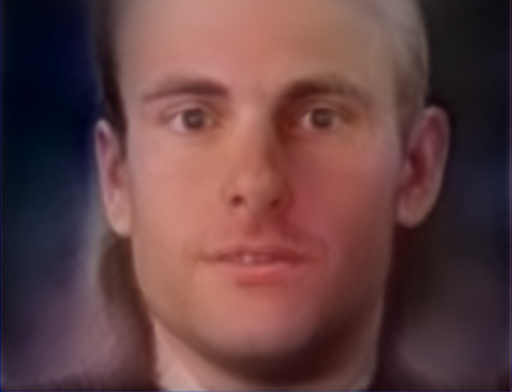}  & 
        \includegraphics[width=\x\linewidth]{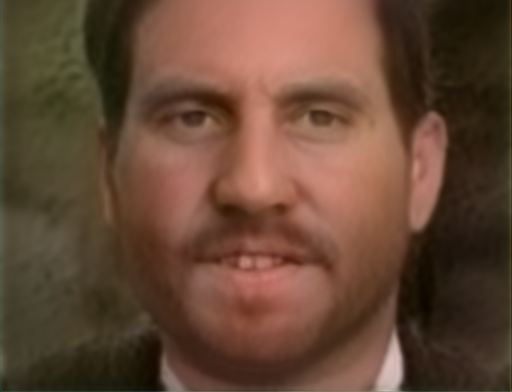}
        \\
        %\vspace{0.005\linewidth}
        \\
        \raisebox{0.7cm}{Ours} & 
        \includegraphics[width=\x\linewidth]{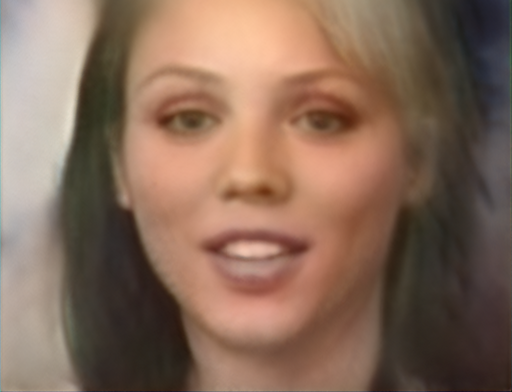} & 
        \includegraphics[width=\x\linewidth]{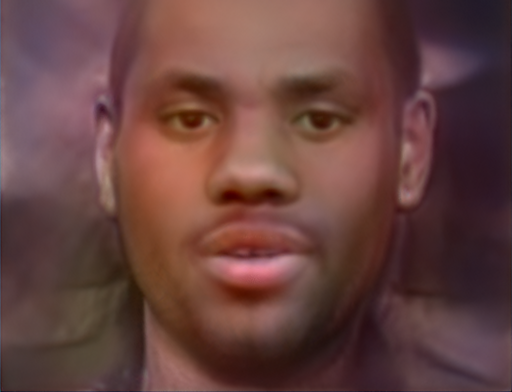} & 
        \includegraphics[width=\x\linewidth]{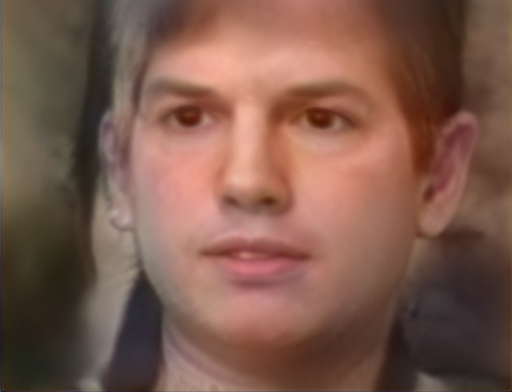} &
        \includegraphics[width=\x\linewidth]{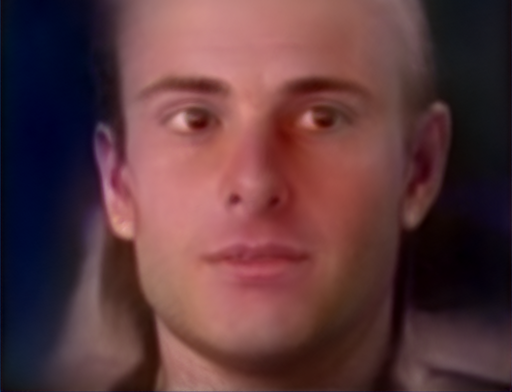}  & 
        \includegraphics[width=\x\linewidth]{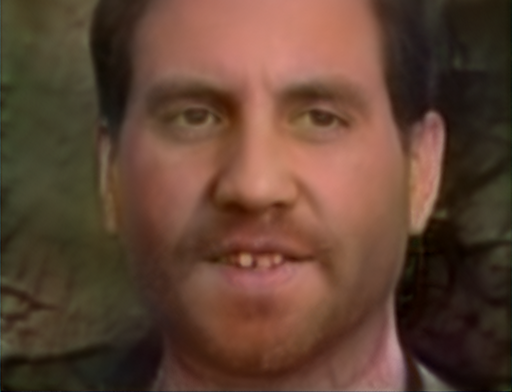}
        \\
        % \vspace{0.01\textwidth} \\
        % \raisebox{0.9cm}{Method} & 
        % \raisebox{0.9cm}{Identity 1} & 
        % \raisebox{0.9cm}{Identity 2} & 
        % \raisebox{0.9cm}{Identity 3} & 
        % \raisebox{0.9cm}{Identity 4} & 
        % \raisebox{0.9cm}{Identity 5} \\
    \end{tabular}
    \vspace*{1mm}
    \caption[Pose and expression transfer comparison]{Pose and expression transfer comparison. The top two rows represent the input: source and target images. The next row shows the results. The baseline methods,  pSp and e4e inversion. The three variants of our method, Ours-Gen with optimized generator weights, Ours-Cos with CosFace loss, and Ours as our best model. %The Top box depicts the identity input image along with their inversion. The bottom grid shows the transfer of pose, expression, and eye movement to different identities. The identities are preserved column-wise, and the poses and expressions are preserved row-wise.}
    }
    \label{fig:comparison}
    \vspace*{-1mm}
\end{figure*}

%popsat varianty GEN, COS
In Fig.~\ref{fig:comparison}, we compare the results of the baseline method with several variants of our proposed method.  The baseline method does not use the target image, but rather a frontal representation with an average expression across the video of the identity, as explained in Sec.~\ref{sec:eval-method}. The figure shows that the baseline methods have trouble preserving the identity of the target person and several visual artifacts are present. Some expressions are transferred relatively faithfully. However, it can happen that the average expression in one video is not the same as in the other, and then the expressions are not transferred correctly. This can be seen in the second and last columns of the Fig.~\ref{fig:comparison}. Our best model represents eye movement better than other variants while also generating more realistic images.

\paragraph{Expression transfer to synthetic faces.}
% Figure \ref{fig:random} shows the expression and pose transfer performed to randomly generated identities via StyleGAN. 
Our method allows for transferring pose and expression onto a randomly generated identities via StyleGAN.
We sample a random latent code \textbf{z} from the Gaussian distribution, which is then mapped by StyleGAN mapping network to \textbf{w} \( \in \mathcal{W}\). To obtain a valid identity latent code for our network, we first generate an image using StyleGAN with \textbf{w} and then invert it using ReStyle. This is due to the fact that ReStyle encodes the images into a specific subspace of StyleGAN's latent space and our model is trained to operate in this subspace. Feeding \textbf{w} directly into our mapping network $M$ often results in certain artifacts. In this way, we can efficiently generate images of random identities with a specific pose and expressions given an example.

\subsection{Quantitative evaluation}\label{sec:eval-quantitative}
We evaluate the proposed method on pose and expression transfer fidelity, as well as on identity preservation. We then compare the results with the baseline methods and other variants of our method. The evaluation is done on the test split of the VoxCeleb2 dataset \cite{voxceleb2} that contains 120 different identities. %The test set is pre-processed in the same way as the training set. 
Our evaluation focuses on a cross-reenactment scenario, i.e.,  the source and target images are from different identities. In particular, for each video in the test set, every frame is one by one taken as the source (driving) image, and a single random frame of another video is taken as the target image (of a different identity) and fed into the model to generate output videos. 
% TODO: explain generated videos 

For pose transfer evaluation, we use a pre-trained CNN estimator~\cite{pose}. The network predicts yaw, pitch, and roll; however, we consider only yaw and pitch since all the pre-processed and generated images have the same roll. The pose error is the mean absolute error of yaw and pitch between the generated images and their corresponding source (driving) images.

For the evaluation of identity preservation, we use the ArcFace~\cite{arcface}. The ID error is the cosine similarity between the generated and the target (identity) frame descriptors. 

To the best of our knowledge, there is no straightforward method for measuring expression transfer fidelity. In theory, the expression independent of identity and pose should be described by activation of Facial Action Units (FAU) ~\cite{ekman1978facial}. However, using a recent state-of-the-art FAU extractor~\cite{opengraph} did not yield meaningful results in our data. The reason is probably that only strong activations are detected and subtle expression changes are not captured at all.  
Therefore, we opted to utilize Facial Landmarks (FL). 
To detect Facial Landmarks we utilize the Dlib library \cite{dlib09} which predicts 68 landmarks on a human face.
We first calculate the aspect ratios of certain facial features following~\cite{soukupova2016eye}. Specifically, we calculate the aspect ratios of both eyes, the mouth, and measure the movement of the eyebrows by calculating the aspect ratio between the eyebrows and the eyes. 
Instead of measuring expression fidelity between single images, we calculate cross-correlation of aspect ratios between (source and generated) videos, to be insensitive to individual facial proportions. %For each frame from a source video, we generate a corresponding output image with a specified identity from a target image. 
In particular, each aspect ratio in the source and generated videos is calculated for all the frames of the videos. This gives us two signals of the same length that are cross-correlated. Finally, the cross-correlations of all aspect ratios are averaged, giving us the final FL statistic.

%Then, we calculate the correlation of each aspect ratio between the source and generated images. Finally, the correlations across all aspect ratios are averaged, giving us the final FL error.

This is a proxy statistic, since it does not capture eyeball movements and does not measure well asymmetric facial expressions, but seems to correlate with subjective quality of facial expression transfer. 
%This metric does not track eye movement and does not measure asymmetric expressions very well (e.g. mouth movement only on one side). Another issue is that people differ in their facial structure. That is why instead of evaluating the expression transfer between single images, we calculate the correlation of aspect ratios between videos.
%For each frame from a source video, we generate a corresponding output image with a specified identity from a target image. Then, we calculate the correlation of each aspect ratio between the source and generated images. Finally, we average these correlations across all aspect ratios to obtain an evaluation of the expression transfer using facial landmarks.

\begin{table}[t]
\resizebox{1.0\linewidth}{!}{\begin{tabular}{|c|ccc|}
 \hline
 Method& Pose(MAE)$\downarrow$&FL(CORR)$\uparrow$&ID(CSIM)$\uparrow$\\
 \hline
 Base pSp  &  8.491   & \textbf{0.656}   &  0.671  \\%even under challenging conditions such as varied ethnicity
 Base e4e   &  8.720   & 0.621   &  0.563  \\
 Ours Gen  &  8.325  & 0.556&  0.760 \\
 Ours Cos & 7.968  & 0.528&  0.762 \\
 Ours  & \textbf{7.673}  & 0.620& \textbf{0.801} \\
 \hline
\end{tabular}}
\vspace*{2mm}
\caption{Quantitative comparison of the baseline method and variants of our method. %The first two rows show the baseline
%method results, first with pSp inversion configuration and second with e4e.
%The last three rows depict the results of our method, the first with the generator
%weight optimization, the second with utilizing the CosFace loss, and the last
% shows the best parameter model. 
Pose error, expression fidelity (measured by facial landmarks), and identity preservation are evaluated.
Symbol $\uparrow$ indicates that larger is better and $\downarrow$ that smaller is better.}
\label{tab:Quantitative_Evaluation}
\vspace*{-3mm}
\end{table}

Tab.~\ref{tab:Quantitative_Evaluation} shows the quantitative comparison of the baseline method and variants of our method on the VoxCeleb2 test set. 
The baseline methods struggle to preserve the identity of the generated person and generate a correct pose, while they are good or comparable in expression transfer fidelity. %The identity preservation is measured by the cosine similarity of ArcFace \cite{arcface} embeddings. 
Our best model achieves ArcFace cosine similarity of $0.8$, which is very good considering that the cosine similarity between the original and inverted images via ReStyle with pSp configuration is $0.83$. Therefore, our method achieves identity preservation close to the maximum possible with ReStyle encoder. 

Our method performs worse with the CosFace loss function (Ours Cos). While the loss function appears to improve image illumination, as reported by~\cite{drobyshev2022megaportraits}, it significantly slowed training and hindered expression and eye movement transfer. The variant with (Ours Gen) optimized generator weights produces overall inferior output compared to the default model, where the generator is fixed. The generated images suffer from unpleasant artifacts while also having a less realistic color scheme. This is probably a consequence of overfitting. 

\paragraph*{Computational demands.}
The speed of inference is very important in practical applications. Our method needs to invert the identity image via ReStyle, which takes approximately half a second on a modern GPU. Then it can generate up to 20 images per second with that identity, given all the images are already aligned. On the other hand, the baseline method requires the inversion of all the images from the source video and target video but then can generate up to 50 images per second. Given two short 5-sec videos with 24 frames per second, which are typical for the VoxCeleb2 dataset, our method generates the entire video in less than 6 secs, whereas the baseline method would require a little over 2 mins.

\section{Conclusions} \label{sec:conclusion}
We presented a method for transferring the pose and expression of a
source face image to a target face image while preserving the identity of the
target face. The proposed method is self-supervised and does not require
labeled data. %Additionally, it fully relies on neural rendering in a one-shot
%setting without using a 3D graphics model of the human face.
We reviewed the existing methods and proposed a new one that is based on
the StyleGAN generator. We extensively evaluated our method on pose and
expression transfer fidelity as well as on identity preservation. We compare
our method to the baseline that utilizes the arithmetic property of StyleGANs
latent space.
We showed that our model transfers pose, expression, and even eye movement under challenging
conditions such as different ethnicity, gender, pose, or illumination between the source and target images. Our method can be used to generate images of random identities with controllable pose and facial expressions by coupling our model with the StyleGAN generator. The inference runs in close to real-time; thus, it is practically usable to generate videos having a driving video and a single still image of a target face. 

The limitation is that certain expressions are not transferred faithfully. For instance, problematic are fully closed eyes, which is probably due to the difficulty of StyleGAN in producing such images. Face images with eyes completely closed were probably not often seen when StyleGAN was trained. The remedy could be a fine-tuning of the generator on problematic images and a certain regularization of the loss function. 

We will make the code and the trained model publicly available. 

\vspace*{-3mm}
\paragraph{Acknowledgement} The research was supported by project SGS23/173/OHK3/3T/13.

{\small
\bibliographystyle{ieee}
\bibliography{cvww_template}
}

\end{document}